\documentclass[runningheads]{llncs}
\usepackage[T1]{fontenc}
\usepackage{todonotes} % for comments

\usepackage{graphicx,verbatim}
\usepackage{hyperref}
\begin{document}
\title{Improving Autism Detection with Multimodal Behavioral Analysis}
%
\begin{comment}  %% Removed for anonymized MICCAI 2025 submission
\author{First Author\inst{1}\orcidID{0000-1111-2222-3333} \and
Second Author\inst{2,3}\orcidID{1111-2222-3333-4444} \and
Third Author\inst{3}\orcidID{2222--3333-4444-5555}}
%
\authorrunning{F. Author et al.}
\institute{Princeton University, Princeton NJ 08544, USA \and
Springer Heidelberg, Tiergartenstr. 17, 69121 Heidelberg, Germany
\email{lncs@springer.com}\\
\url{http://www.springer.com/gp/computer-science/lncs} \and
ABC Institute, Rupert-Karls-University Heidelberg, Heidelberg, Germany\\
\email{\{abc,lncs\}@uni-heidelberg.de}}

\end{comment}

\author{William Saakyan\inst{1} \and
Matthias Norden\inst{1} \and
Lola Eversmann\inst{2} \and
Simon Kirsch\inst{3} \and
Muyu Lin\inst{2} \and
Simon Guendelman\inst{2} \and
Isabel Dziobek\inst{2} \and
Hanna Drimalla\inst{1}
}
% index{Last Name, First Name}
%
\authorrunning{W. Saakyan et al.}
\institute{Center for Cognitive Interaction Technology (CITEC), Bielefeld University \email{Corresponding: drimalla@techfak.uni-bielefeld.de}
\and
Institute of Psychology, Humboldt University of Berlin \and
Department of Psychiatry and Psychotherapy, Medical Center-University of Freiburg
}

\maketitle              % typeset the header of the contribution
\begin{abstract}
Due to the complex and resource-intensive nature of diagnosing Autism Spectrum Condition (ASC), several computer-aided diagnostic support methods have been proposed to detect autism by analyzing behavioral cues in patient video data. While these models show promising results on some datasets, they struggle with poor gaze feature performance and lack of real-world generalizability. To tackle these challenges, we analyze a standardized video dataset comprising 168 participants with ASC (46\% female) and 157 non-autistic participants (46\% female), making it, to our knowledge, the largest and most balanced dataset available. We conduct a multimodal analysis of facial expressions, voice prosody, head motion, heart rate variability (HRV), and gaze behavior. To address the limitations of prior gaze models, we introduce novel statistical descriptors that quantify variability in eye gaze angles, improving gaze-based classification accuracy from 64\% to 69\% and aligning computational findings with clinical research on gaze aversion in ASC. Using late fusion, we achieve a classification accuracy of 74\%, demonstrating the effectiveness of integrating behavioral markers across multiple modalities. Our findings highlight the potential for scalable, video-based screening tools to support autism assessment. To facilitate reproducibility, we share our code on GitHub: \href{https://github.com/mbp-lab/miccai25_sit_autism_classification}{https://github.com/mbp-lab/miccai25\_sit\_autism\_classification}.


\end{abstract}

\section{Introduction}
\label{sec:introduction}

Impairments in social communication are a key characteristic of Autism Spectrum Condition (ASC), affecting an individual’s ability to interpret and respond to non-verbal cues such as facial expressions, eye contact, and vocal tone \cite{american_psychiatric_association_diagnostic_2013}. The clinical diagnosis of ASC currently relies on subjective assessments, including standardized instruments (e.g., Autism Diagnostic Observation Schedule (ADOS), the Autism Diagnostic Interview-Revised (ADI-R) \cite{Boelte2005}), which are time consuming and depend on the availability of experts. This often leads to long waiting times and potential misdiagnoses, particularly in adults and females who may present atypical symptoms \cite{McQuaid2022,Kirkovski2013}. This highlights the urgent need for objective, scalable, and accessible tools to support autism screening and behavioral assessment.  

Recent advancements in machine learning (ML) have enabled the automated analysis of non-verbal behaviors to assist with ASC detection via video-based approaches \cite{deBelen2020}. Studies have analyzed facial expressions \cite{Briot2021}, gaze behavior \cite{Baron-Cohen2001,Klin2002}, and prosodic features \cite{Bone2015a} to identify ASC-related patterns. These approaches can complement existing diagnostic frameworks and facilitate early detection, remote screening, and large-scale behavioral studies. However, despite this progress, significant challenges remain. 

Most datasets used for computer-aided ASC assessment include children, aiming for early diagnosis \cite{Li2023,Billing2020}. However, ASC is often undiagnosed until adulthood, particularly in females and individuals with milder traits \cite{Huang2020}. Datasets with adult populations remain scarce, and existing ones, such as \cite{Georgescu2019}, have limited sample sizes (58 individuals) and feature sets. Moreover, most datasets are collected in highly controlled laboratory environments, limiting model robustness in naturalistic settings. Additionally, many data types, such as neuroimaging, molecular, and genetic data, require expensive equipment, creating development and deployment challenges of AI-based tools for ASC assessment.

Social interaction is multimodal, involving gaze, facial expressions, vocal prosody, and body movements \cite{Nota2021}. While multimodal ML models have improved classification accuracy \cite{Sarwani2024,Derbali2023}, many over-rely on facial features, underutilizing other modalities, such as gaze behavior.

Atypical gaze is a well-established ASC marker \cite{Minissi2022,Jones2013}. Reduced eye contact and increased gaze aversion have been observed across lab-based and naturalistic settings \cite{Papagiannopoulou2014}. While some studies have successfully used eye-tracking devices to analyze gaze behavior in ASC \cite{Islam2025}, these approaches often rely on specialized hardware or controlled experimental tasks rather than naturalistic social interactions. Webcam-based interaction analyses, such as \cite{Drimalla2020,Saakyan2023}, have reported poor performance for gaze-based ASC classification, likely due to simplistic descriptors that fail to capture gaze behavior in relation to social stimuli.

Several studies suggest heart rate variability (HRV) as a biomarker for ASC due to its link to autonomic nervous system function. Research indicates that individuals with ASC often exhibit autonomic dysregulation, characterized by both hyperarousal and hypoarousal states at rest, which may impact their ability to engage with social environments and regulate sensory input\cite{Arora2021}. \cite{Thapa2019,Thapa2021} found significantly lower resting-state HRV in both adults and children with ASC. \cite{Frasch2021} applied machine learning to HRV data, achieving an AUC of 0.89 for ASC classification. While HRV shows promise as a non-invasive biomarker, its role in ASC remains complex, with findings influenced by measurement context, heterogeneity in study methodologies, and individual variability\cite{Arora2021}.

To address the challenges of subjective ASC diagnosis, limited dataset availability for adults, and the underutilization of key social interaction markers, we improve computer-aided autism detection by evaluating multimodal behavioral markers using the largest adult social interaction dataset to date. To establish this dataset, we used the Simulated Interaction Task (SIT) \cite{Drimalla2019,Drimalla2020} which elicits standardized, yet naturalistic social behaviors by presenting participants with a video-recorded conversational partner. Unlike prior studies using lab-based, child-focused datasets, we included adult participants in both clinical and home settings, providing a more valid dataset. To our knowledge, this is the largest and most balanced dataset available.

Our approach introduces enhanced gaze behavior descriptors alongside facial expression, head movement, voice prosody, and heart rate variability (HRV) features to improve diagnostic performance. We systematically evaluate unimodal and multimodal feature combinations to identify the most informative digital biomarkers for video-based ASC detection. By combining behavioral analysis with computational modeling, our work helps bridge the gap between computer aided ASC detection and clinical practice, contributing to the development of scalable, non-invasive screening tools for diverse populations.

\section{Methods}
\label{sec:methods}

\subsection{Dataset}
\label{subsec:dataset}
We collected a large-scale and diverse video dataset including 168 participants with ASC (46\% female) and 157 controls (46\% female), recorded in clinical ($n=254$) and home settings ($n = 71$)  using the Simulated Interaction Task (SIT) paradigm \cite{Drimalla2020}. 
Lab study participants were recruited for a larger research project (number DRKS00017817). Inclusion criteria: age 18–65, IQ $\geq$80, fluency in German, and stable or no pharmacotherapy. Exclusion criteria included psychiatric comorbidities of schizophrenia, psychosis, severe depression, acute manic episodes within bipolar disorder, and acute suicidality. ASC diagnoses were confirmed by licensed clinicians using ICD-10 criteria, while the non-autistic participants reported no psychiatric diagnoses. 
Home-study participants were recruited via clinics, therapy groups, and online postings. ASC diagnoses were confirmed via medical records. As we were specifically interested in social interaction behavior related to ASC, we excluded participants reporting comorbid conditions that could impact social interaction (e.g., Social Anxiety Disorder, Depression).
The studies were approved by the respective ethical committee of Humboldt University of Berlin and the Medical Center-University of Freiburg (approval 20-1144\_3 and 2021-20, \url{https://doi.org/10.1186/s13063-021-05205-9}).

\begin{table}[h]
    \centering
    \caption{Gender (male, female, diverse) and age distributions by setting and group.}
    \begin{tabular}{|l|ccc|cc|cc|cc|c|}
        \hline
        \textbf{Study Setting} & \multicolumn{3}{c|}{\textbf{ASC}} & \multicolumn{2}{c|}{\textbf{Non-ASC}} & \multicolumn{2}{c|}{\textbf{ASC Age}} & \multicolumn{2}{c|}{\textbf{Non-ASC Age}} & \textbf{Total} \\
        \cline{2-10}
        & \textbf{M} & \textbf{F} & \textbf{D} & \textbf{M} & \textbf{F} & \textbf{Median} & \textbf{Range} & \textbf{Median} & \textbf{Range} & \\
        \hline
        Home & 13 & 13 & - & 24 & 21 & 36 & 18-57 & 27 & 18-60 & 71 \\
        Lab  & 74 & 65 & 3 & 60 & 52 & 34 & 18-63 & 35 & 18-64 & 254 \\
        \hline
        \textbf{Total} & 87 & 78 & 3 & 84 & 73 & - & - & - & - & 325 \\
        \hline
    \end{tabular}
    \label{tab:demographics_distribution}
\end{table}

\subsubsection{Procedure} 
Participants completed the SIT application on a computer in lab or home settings. The fully automated procedure began with head positioning for facial landmark calibration. The conversation scenario included three phases: Meal preparation ("Neutral"), Favorite foods ("Joy"), and Disliked foods ("Disgust"). Each phase consisted of two interactions: the actress speaking while the participant listened, followed by the participant speaking while the actress displayed empathic listening.

\subsection{Preprocessing}
To ensure consistency with previous research \cite{Saakyan2023}, we applied the same video preprocessing pipeline: For videos with frame rates between 15 and 25 FPS, we applied linear interpolation to achieve a uniform 30 FPS frame rate. Videos with frame rates below 15 FPS were excluded from analysis due to motion artifacts. 

\subsection{Features}
We extracted non-verbal features from following modalities: facial expressions, gaze, head movement, paralinguistics, and HRV, using open-source libraries.

Feature extraction was performed across six interaction phases, corresponding to three emotion-specific segments (neutral, joy, disgust) and two speaking roles (participant speaking and participant listening). Audio features were extracted only from participant-speaking segments, while HRV signals were extracted primarily during listening segments to minimize motion artifacts.

\subsubsection{Visual}
Visual features were extracted using OpenFace 2.2 \cite{Baltrusaitis2018}, which detects Action Units based on the Facial Action Coding System, head position (rotation angles: pitch, yaw, roll), and gaze direction (angle $x$ and $y$). Frames with a detection confidence below 75\% were excluded. Participants with more than 10\% invalid frames were removed (five in total).

\textbf{Facial expression:} OpenFace computes 18 AUs, each with a presence score (binary) and an intensity score (0–5). We calculated the mean, median, and standard deviation of the AU intensities in each of the six interaction phases, as well as the mean AU presence. Additionally, we computed the AU onset frequency, which is the number of activation onsets (transitions from 0 to 1) per phase.

\textbf{Head movement} metrics, including velocity, acceleration, and stability durations, were calculated as in previous work \cite{Saakyan2023}, along with the mean and standard deviation of yaw and roll rotation angles. Additionally, we included the pitch angle and, to mitigate potential gender and height biases, normalized the values based on each participant's median pitch angle. To provide a robust measure of spread, we computed the interquartile range (IQR) alongside other summary statistics. Furthermore, we implemented a  nod detection algorithm that scans the pitch trajectories for characteristic downward-upward sequences occurring within 1.5-second windows.

\textbf{Gaze behavior:} Following \cite{Saakyan2023}, we computed gaze movement velocity, acceleration, saccade amplitude, and fixation duration, as well as the mean and standard deviation of the $x$ angle. Additionally, we included statistics for the $y$ angle, which was normalized relative to each participant’s median $y$ angle to mitigate height-related biases. To move beyond angle-based gaze descriptors and capture socially meaningful gaze behavior, we implemented a geometric transformation that maps gaze angles from the webcam coordinate system to a screen-centered coordinate space. For lab participants, this transformation involved measuring camera-to-screen distance, eye-to-screen distance, and screen resolution to determine gaze position relative to the screen. For home participants, gaze position was approximated. From these transformed coordinates, we computed: Mean screen fixation time (proportion of frames with gaze directed toward the screen area containing the actress's face), Number of off-screen fixations (frequency of gaze shifts beyond screen boundaries, indicating gaze aversion), Euclidean distance from the screen center (mean, standard deviation, skewness, kurtosis, minimum, and maximum distance from the actress's face).

\subsubsection{Audio}
Paralinguistic features from the voice were extracted using OpenSmile 3.0 \cite{Eyben2010} with the eGeMAPSv02 feature set \cite{Yang2022a}for each participant-speaking phase. Extracted features include pitch (mean, variance), intensity, shimmer, jitter, harmonic-to-noise ratio (HNR), and formant frequencies.

\subsubsection{Heart rate}
HR features were extracted using the rPPG Toolbox \cite{Liua}, applying face cropping and the Plane-Orthogonal-to-Skin (POS) method to estimate remote photoplethysmographic (rPPG) signals. HeartPy was used to extract HR features from the rPPG signals. HR signals were extracted during listening phases to minimize motion artifacts. The rPPG Toolbox failed to extract certain feature values for 18 participants. We imputed the missing values using the median from the training set. Features included mean HR, HR variability (HRV), root mean square of successive differences (RMSSD), and low-frequency to high-frequency power ratio (LF/HF). 
%HR variability may reflect differences in autonomic arousal during social interactions, with ASC individuals often exhibiting reduced parasympathetic reactivity \cite{autism_hr_study}.

\subsection{Classification and analysis}

We used XGBoost, a gradient boosting decision tree model, which shows its effectiveness in handling structured tabular data\cite{Shwartz-Ziv2022}. Model parameters were set to default XGBoost settings, to ensure comparability with \cite{Saakyan2023}: gbtree as booster, learning rate of 0.3, maximum depth of tree was 6, XGBoost version 2.0.3. The classification task aimed to differentiate between individuals with ASC and non-autistic individuals using the extracted multimodal features. We calculated performance metrics, including accuracy, precision and recall and compared unimodal models, where each modality was evaluated separately, with multimodal fusion approaches. Early fusion involved concatenating all extracted features into a single feature vector. In the late fusion approach, we combined the probability scores from unimodal models and classified them using logistic regression, incorporating polynomial features (degree 2) to capture non-linear relationships and interactions. All analyses reported in this paper were evaluated using a participant-based Leave-One-Out Cross-Validation \cite{Webb2010} approach.

To gain deeper insights beyond overall model performance, we conducted several follow-up analyses. First, we applied SHapley Additive exPlanations (SHAP) to identify the key features contributing to ASC classification. Second, we performed a statistical analysis of misclassifications to determine whether errors were influenced by dataset source (home vs. lab), participant gender, or Autism Spectrum Quotient (AQ) score. Lastly, we assessed the contribution of each modality by systematically excluding them in late fusion setting.

\section{Results and Discussion}
\label{sec:results}

\subsection{Model performance}
\begin{figure}
    \centering
    \includegraphics[width=0.43\textwidth]{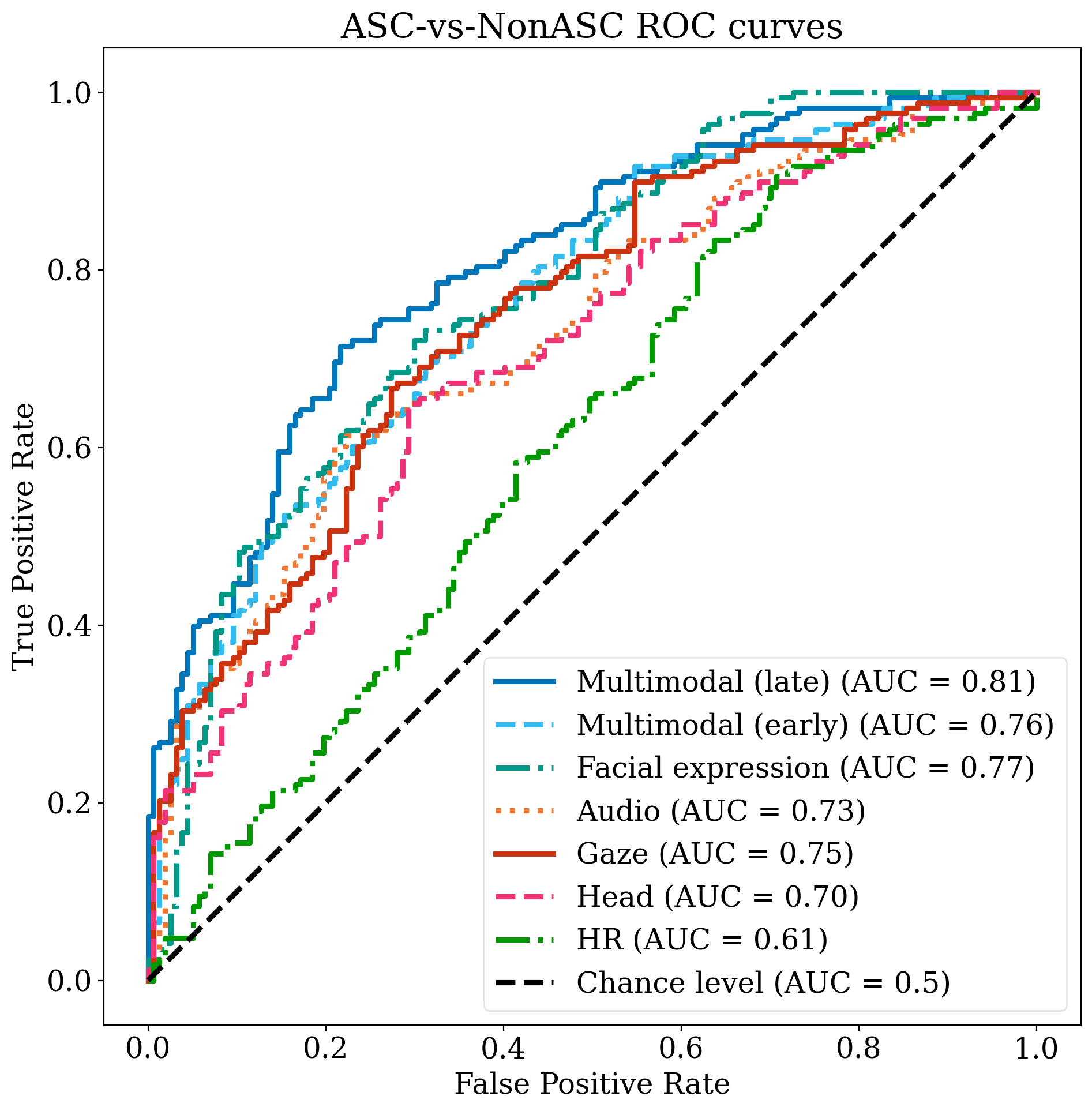}
    \includegraphics[width=0.55\textwidth]{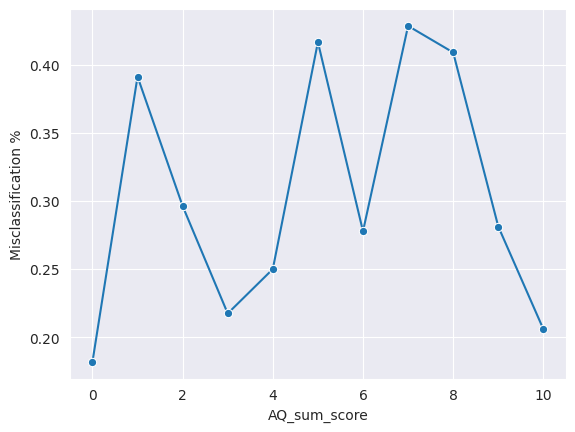}
    \caption{\textbf{Left}: ROC \textbf{Right}: Misclassifications of late fusion model across AQ scores}
    \label{fig:roc}
\end{figure}

We conducted a comprehensive uni- and multimodal video analysis using machine learning on a large, gender- and context-balanced dataset, including 325 participants. Table \ref{tab:classification} presents classification performances, while Figure \ref{fig:roc} illustrates the corresponding ROC curves. Our refinement of gaze features led to the most significant improvement, increasing accuracy by 5 percentage points compared to previous works using a similar approach. Additionally, enhanced feature engineering for the head modality resulted in a 2-point accuracy gain. Multimodal late fusion achieved the highest classification accuracy of 74\%, outperforming the feature set of prior work by 6 percentage points.

\begin{table}
\caption{ASC classification performance for different modalities using the previous\cite{Saakyan2023} and current feature sets. Bold values indicate the highest accuracy per modality.}
    \begin{tabular}{|l|c|c|c||c|c|c|}
        \hline
        \textbf{Modality} & \multicolumn{3}{c||}{\textbf{Previous Feature Set}} & \multicolumn{3}{c|}{\textbf{Current Feature Set}} \\
        \cline{2-7}
        & \textbf{Accuracy} & \textbf{Precision} & \textbf{Recall} & \textbf{Accuracy} & \textbf{Precision} & \textbf{Recall} \\
        \hline
        Multimodal (late)  & 0.68 & 0.69 & 0.68 & \textbf{0.74} & 0.76 & 0.74 \\
        Multimodal (early) & 0.67 & 0.67 & 0.71 & \textbf{0.68} & 0.68 & 0.71 \\
        Facial expression            & \textbf{0.70} & 0.70 & 0.73 & \textbf{0.70} & 0.71 & 0.73 \\
        Audio              & \textbf{0.66} & 0.67 & 0.66 & \textbf{0.66} & 0.67 & 0.66 \\
        Gaze               & 0.64 & 0.66 & 0.64 & \textbf{0.69} & 0.70 & 0.68 \\
        Head               & 0.63 & 0.64 & 0.66 & \textbf{0.65} & 0.66 & 0.68 \\
        HR                 & -    & -    & -    & \textbf{0.57} & 0.58 & 0.63 \\
        \hline
    \end{tabular}
    \label{tab:classification}
\end{table}

\subsection{Gaze Behavior Analysis}
\begin{figure}
    \centering
    \includegraphics[width=0.49\textwidth]{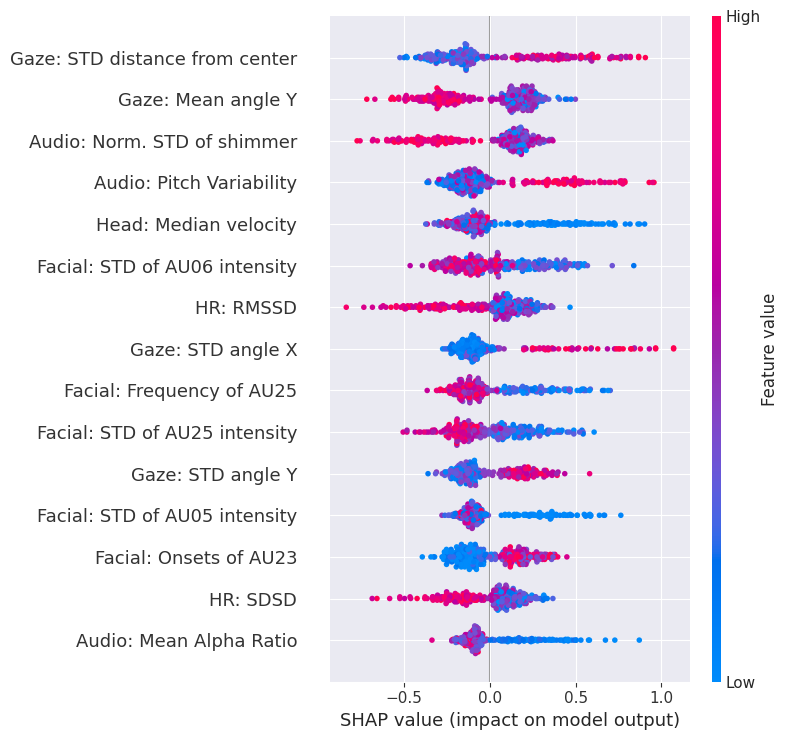}
    \includegraphics[width=0.49\textwidth,trim={0 0.31cm 0cm 0.3cm},clip]{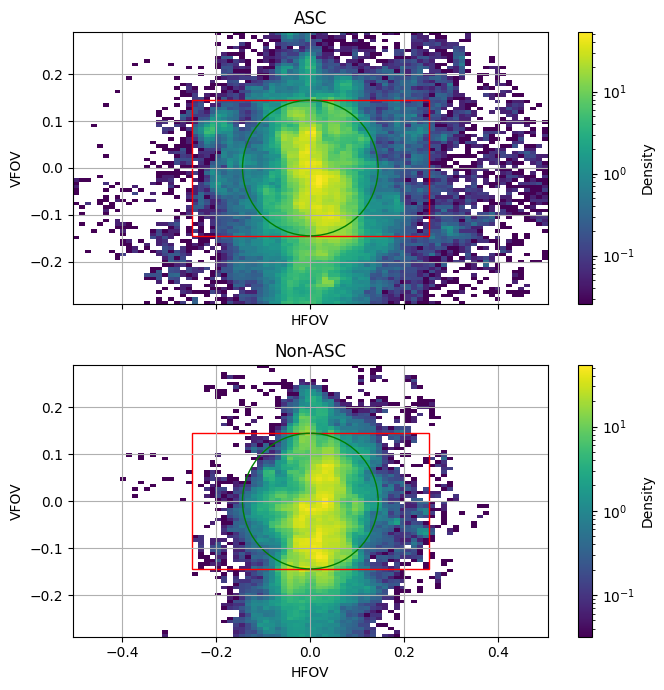}
    \caption{\textbf{Left}: SHAP values for the multimodal early fusion model. Each dot represents a feature's impact on an individual prediction. \textbf{Right}: Eye gaze angle projections on the screen surface for ASC and non-autistic participants during the passive parts. The red rectangle represents the screen area, where the actress was displayed. Vertical and horizontal axes represent fields of view.}
    \label{fig:shap_aversion}
\end{figure}

Prior computational studies have reported inconsistent results regarding gaze behavior in ASC \cite{Drimalla2020,Saakyan2023}, despite psychological research suggesting its diagnostic relevance \cite{Jones2013,Klin2002}. These discrepancies are likely due to simplistic angle-based descriptors that fail to fully capture gaze aversion patterns. In our analysis, gaze behavior exhibited the largest performance gain among unimodal models, demonstrating the effectiveness of our new screen-centered descriptors in capturing gaze aversion. Figure \ref{fig:shap_aversion} presents the SHAP analysis, identifying gaze distance variability from the screen center as the most influential feature. On the right side, we visualize eye gaze angles projected onto the screen surface, further illustrating these differences. Our results confirm that ASC participants exhibit significantly greater gaze variability than non-autistic individuals (\(+63.5\%\)), particularly in the Disgust phase (\( p = 0.003, r = -0.215 \)).

\subsection{Misclassification Analysis}
We analyzed \textbf{misclassifications} for potential biases. Chi-square tests revealed no significant differences in classification errors based on gender or recording environment (\( p \geq 0.05 \)). There was no evidence of an influence of gender or recording setting (lab vs. home) on model performance, suggesting a degree of generalizability and potential real-world applicability.

Given the overlap between ASC and non-autistic individuals in middle AQ score ranges, we expected more frequent misclassification for participants with intermediate scores. However, a Mann–Whitney U test comparing the AQ score distributions of correctly and incorrectly classified participants revealed no significant difference (\( p \geq 0.05 \)). Figure \ref{fig:roc} (right) shows the distribution of late-fusion model misclassifications across AQ scores.

\subsection{Contribution of each modality}
We evaluated each modality’s contribution by measuring performance drops after removal. Excluding eye gaze or facial expressions led to the largest accuracy reduction (-4 percentage points), highlighting their importance in the model’s predictions. This aligns with prior research emphasizing atypical gaze behavior \cite{Jones2013,Baron-Cohen2001,Klin2002} and reduced facial expressivity \cite{Briot2021} as core characteristics of ASC. Despite its low standalone accuracy, removing HR variability still decreased performance by -2 percentage points, supporting evidence that autonomic regulation differs in ASC individuals \cite{Bone2015a}. A similar drop for audio features reinforces the relevance of prosodic differences for ASC detection \cite{Bone2015a}. In contrast, removing head pose data had no effect, suggesting minimal contribution.

\subsection{Conclusion}
This study contributes to computer-aided ASC assessment by providing a large-scale, adult-focused dataset and an in-depth analysis of multimodal behavioral markers in standardized interactions. Our results highlight the value of incorporating more precise gaze features, demonstrating that improved gaze data enhances ASC classification accuracy.

While our study focused on aggregated behavioral summaries for distinct interaction phases, future work should explore time-dependent social interaction dynamics using recurrent neural networks (LSTMs) or transformer-based architectures to capture fine-grained temporal fluctuations in gaze shifts, facial expressivity, and vocal prosody. Improving the reliability of webcam-based HR extraction, including advancements in rPPG techniques, is another promising avenue. %Finally, ASC detection models should not focus solely on accuracy but also on capturing diagnostically meaningful behaviors.
Furthermore, while our model identifies behavioral differences between individuals with and without ASC, clinical classification involves differentiating ASC from other conditions with overlapping social impairments, such as social anxiety or personality disorders. Future work should evaluate how well multimodal behavioral markers disentangle ASC-specific traits from social difficulties of other clinical conditions.
Nevertheless, our findings demonstrate the potential of multimodal behavioral analysis for ASC detection.

\begin{credits}
\subsubsection{\ackname} Research project was funded by the Deutsche Forschungsgemeinschaft (DFG, German Research Foundation) under Germany's Excellence Strategy (EXC-2049 – 390688087). The DFG had no role in study design, data collection, analysis and interpretation, the decision to write and submit the article for publication.

\subsubsection{\discintname}
The authors have no competing interests to declare that are relevant to the content of this article. 
\end{credits}

\bibliographystyle{splncs04}
\bibliography{MICCAI2025}

\end{document}